\documentclass[11pt,twocolumn]{article}
\usepackage{graphicx}
\usepackage{amsmath}
\usepackage{amssymb}
\usepackage{mathtools}
\usepackage{subfigure}
\usepackage{units}
\usepackage[square,sort,comma,numbers]{natbib}

\begin{document}
%

\title{ConceptVision:\\
A Flexible Scene Classification Framework
}

%
%
%
%
%

%
\author{
%
%
Ahmet Iscen\\
       {Computer Engineering Department}\\
       {Bilkent University}\\
       {ahmet.iscen@bilkent.edu.tr}
       \and
Eren Golge\\
       {Computer Engineering Department}\\
       {Bilkent University}\\
       {eren.golge@bilkent.edu.tr}
\and
Ilker Sarac\\
       {Computer Engineering Department}\\
       {Bilkent University}\\
       {eren.golge@bilkent.edu.tr}
\and
 Pinar Duygulu\\
       {Computer Engineering Department}\\
       {Bilkent University}\\
       {duygulu@cs.bilkent.edu.tr}
}
\date{}

\maketitle
\begin{abstract}
\vspace{-0.5cm}
We introduce ConceptVision, a method that aims for high accuracy in categorizing large number of scenes, while keeping the model relatively simpler and efficient for scalability. The
proposed method combines the advantages of both low-level representations and high-level semantic categories, and eliminates the distinctions between different levels through the definition of concepts. The proposed framework encodes the perspectives brought through different concepts by considering them in concept groups. Different perspectives are ensembled for the final decision. Extensive experiments are carried out on benchmark datasets to test the effects of different concepts, and methods used to ensemble. Comparisons with state-of-the-art studies show that we can achieve better results with incorporation of concepts in different levels with different perspectives.
\end{abstract}



\vspace{-0.3cm}
\section{Introduction}
\label{sec:sec1}
With the advancement of digital cameras and smart phones,
billions of images have been stored in personal collections and
shared in social networks. Due to the limitation and subjectivity of
tags associated with images, methods that can categorize images
based on visual information are required to manage such a huge
volume of data. On the other hand, it is still a challenge to classify
images when the number and the variety of images are large.

\begin{figure}
\centering
\subfigure[]{\includegraphics[width=0.45\linewidth]{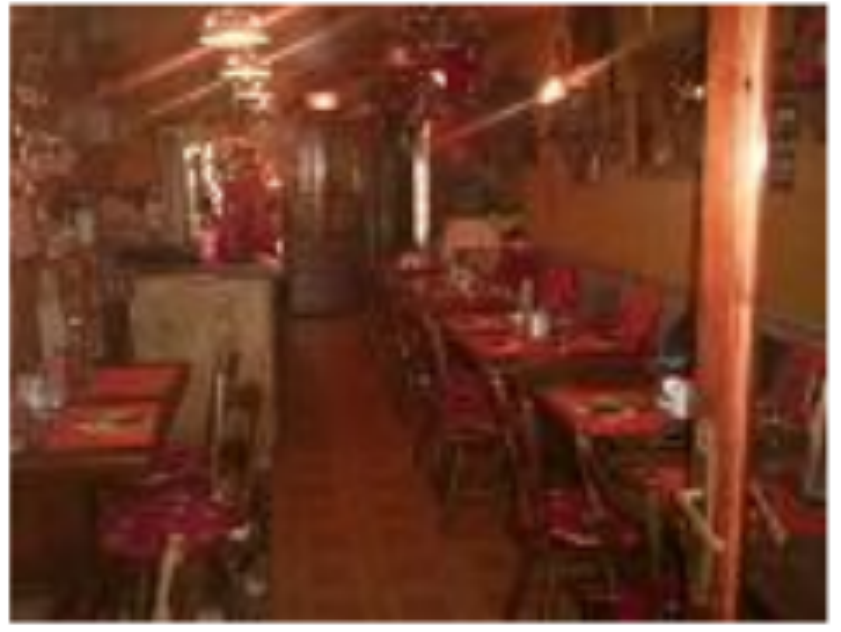}}        
\subfigure[]{\includegraphics[width=0.45\linewidth]{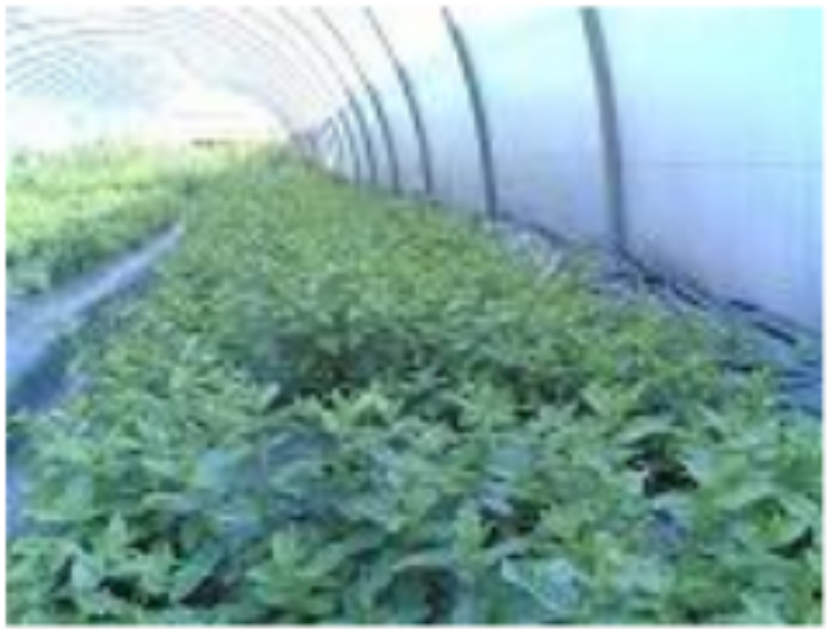}}    
    \vspace{-0.3cm}  		
\caption{While image (a) can be described
through semantic categories such as chair, table, floor, wood, image (b) can only be described with the help of low-level features such as color, texture.}
\label{fig:fig1}
\vspace{-0.7cm}
\end{figure}  

Scene categorization has been attacked by many studies in
computer vision and multimedia literature. Low-level features
extracted over the entire image are commonly used to classify
scenes, such as for indoor versus outdoor, or city versus landscape categorization \cite{ref1,ref2,ref3,ref4}. Recently, Oliva and Torralba proposed to represent scenes with a set of holistic spatial scene properties, which are referred to as Spatial Envelope, such as degree of naturalness, openness, roughness, etc \cite{ref5}.

Histogram of quantized local descriptors, usually referred to as
visual words, is shown as a simple but effective way for
representation of scenes \cite{ref6,ref7,ref8} as an alternative to global
features. Spatial pyramid matching is proposed in \cite{ref9} where
histograms are computed for different levels of image partitions.

On the other direction, with the idea that semantics is not
sufficiently captured by low-level features, object detector
responses have been used as high-level representations \cite{ref10,ref11}.
While the number of objects could reach to hundreds -thanks to
recent availability of good detectors that can be generalized for a
large variety of objects \cite{ref12}-, the main drawback of object-based
approaches is the requirement for manual labeling to train the
object models. Moreover, it may be difficult to describe some
image through specific objects.

Recently, use of attributes as mid-level representations has also
gained attention \cite{ref13,ref14,ref15}. Objects are described through a set of
attributes that are shared between object categories, such as object
parts (wheels, legs) or adjectives (round, striped). However, these
methods also heavily depend on training to model human-defined
attributes.

These problems bring us back to the discussion of semantic
concepts for classification and retrieval of images and videos
\cite{ref16,ref17}.
How can we describe scenes through a set of
intermediate representations? Should they correspond to only
semantic categories? How many of those are required to capture
all the details?

Let's consider the images in Figure \ref{fig:fig1}. The object names such as
chair, table, floor, and the attributes such as wood are relatively
easier to come up with in describing the scene. On the other hand,
how much can we describe the image on the right with nouns or
adjectives that can be learned through examples in training? How
much information can be captured in both images through only
low-level features such as the color or edge distribution?

We argue that, while the advantages of mid- or high-level
semantic categories cannot be discarded, because of the high cost
of training -even if we do not consider the question of
definability-, it could be provided only in limited numbers.
Therefore, low-level information, which is less-costly to obtain,
should be taken advantage of. The main question is how can we
melt representations with different characteristics in the same
pool?

In this study, we introduce ConceptVision, in which we use the
term concept for any type of intermediate representation, ranging
from visual words to attributes and objects. We do not restrict
ourselves with only semantic categories that can be described by
humans, but consider also low-level representations. We handle
the variations between different levels of concepts, by putting
them into concept groups. Separate classifiers are trained for each
concept group. The contributions of each concept group to the
final categorization are provided in the form of confidence values
that are ensembled through a set of methods for the final decision.

In the following, first we present our proposed method in Section \ref{sec:sec2}. We describe the datasets used and discuss about the
implementation details in Section \ref{sec:sec3}. Section \ref{sec:sec4} shows our
experimental results with ConceptVision and Section \ref{sec:sec5} provides
comparisons with other methods.
\vspace{-0.3cm}
\section{Our Method}
\label{sec:sec2}

\begin{figure*}[t]
\centering
\subfigure{\includegraphics[width=1\textwidth]{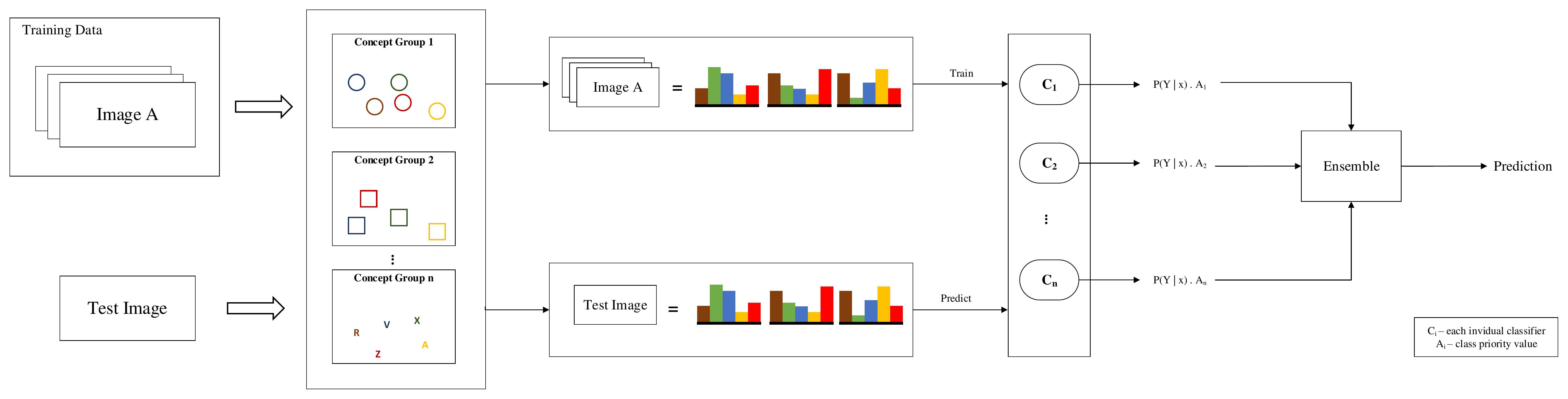}}
    \vspace{-0.7cm}
\caption{A visual representation of ConceptVision. In the training stage, concepts in each concept group are obtained from training
images. Then, an individual classifier is trained for each concept group. Also, for each trained individual classifier, we compute a concept-priority value. In testing stage, a test image, represented by concepts, is fed to the individual classifier of each concept group. Class-confidence values of each classifier incorporated with the concept-priority values, are combined in the ensemble stage to make a final prediction.} 
\label{fig:method_figure}
    \vspace{-0.6cm}
\end{figure*}

As depicted in Figure \ref{fig:method_figure}, ConceptVision brings the ability of using
different levels of descriptors through the definition of concepts
and concept groups. Low-level local or global descriptors could
be quantized to obtain concepts in the form of visual words, and
then concept group can be represented as Bag-of-Words. On the
other hand, each object category could correspond to a concept,
and as a whole the concept group could be represented through a
vector of confidence values of object detectors.

ConceptVision is designed to allow the integration of different
\textit{concept groups} for classification. Note that our definition does
not necessarily require concept groups to have any semantic
meaning; we suppose that, with a capable combination technique,
each concept group can add a different perspective for
classification. Hence, the proposed framework tries to capture the
perspective of each concept group and consider it during the final
classification.

Our classification has two main parts; \textit{individual classification}
and \textit{ensemble of classifiers}. Individual classification is applied to
each concept group separately, and sequentially each
classification result is combined in ensemble of classifiers stage
before making a final prediction. Following sub-sections discuss
these stages in more detail.

\subsection{Individual Classification}
\label{sec:sec21}
To support our hypothesis of trying to examine the different
perspectives of each concept group, we consider each group
independently in the beginning. That is, we assume that the
individual classification performance of a concept group has no
effect on another, and should therefore be treated completely
separately. This also allows us to have an agnostic classification
method that can be used with any type of \textit{concepts}.

In order to implement this idea, we train a separate, individual
probabilistic classifier for each concept group. For a given image
query, the role of each individual classifier is to give the
probability of the image belonging to each class. In
ConceptVision, we use probabilistic SVM as our classifier type.
SVM is a popular classifier that is used widely in any area that
involves machine learning. While the regular SVM outputs a
single prediction class for a given query, a probabilistic SVM
gives class-confidences, the probability of the query belonging to each class.

The \textit{individual classification} phase is simple, and can be
summarized in the following steps:

\begin{enumerate}
\item Collect the extracted concepts for each concept group
using the training set.
\item Train a separate multi-class SVM classifier for each
concept group.
\item During the testing phase, apply concepts extracted from
test set to the corresponding trained probabilistic SVM
model. Output of the model gives the probabilities of
query images belonging to each class.
\end{enumerate}
\vspace{-0.4cm}
\subsection{Ensemble of Classifiers}
\label{sec:sec22}
Ensemble of classifiers is the most important stage of our
framework. After training a separate classifier for each concept
group, we must be able to combine them properly before making a
final decision. Before making any further advancement, we must
first answer a simple question. How do we consider each
individual classifier? Do we consider them equally, or do we
assign more priority to some of them than others? And if latter,
what criteria should we use to assign more priority to an
individual classifier?

We first examine the first option, not assigning any priorities to
individual classifiers and treating them equally. This approach
would simply take outputs of each classifier and combine them
without making any other operations. However, just after a brief
pondering we can intuitionally detect some flaws or misjudgments
with this approach. First of all, how can we guarantee that each
individual classifier will perform well? In fact, we cannot guarantee anything with unknown data, and a single poor-
performing classifier would have the same contribution in the
final decision making process as a well-performing classifier. This
would include many noisy factors for the final prediction making;
therefore we decide to explore the second approach, giving
priorities to each individual classifier.

This brings us to the second question, how do we decide which
individual classifier gets which priority? To answer this question,
we must find a way to have an estimate on how a classifier would
work on general unseen data, so we can assign more weight to
decisions of those that are expected perform well, and less weight
to those that are predicted to perform poorly.

We introduce the notation of \textit{concept-priority value}. \textit{Concept-priority value} is a value which serves as an estimate on how each classifier would perform generally. We find this value by performing a k-fold cross-validation on the training set using each each classifier and assigning the average accuracy value as the \textit{concept-priority value} of the corresponding individual classifier.

Now that we have a generalized estimation for the performance of each individual classifier, we can weigh their outputs accordingly. Probability outputs of each single classifier is multiplied by its \textit{concept-priority value}. After obtaining the weighted class-confidence
probabilities from each classifier, we ensemble them together in
the final step. At the end, the class that obtains the highest value is
selected as the final prediction. Different techniques considered
for the ensembling will be discussed in Section \ref{sec:sec4}.
\vspace{-0.3cm}

\section{Implementation}
\label{sec:sec3}
To demonstrate the ConceptVision idea, it is desired to include
concepts at different levels. In order to eliminate effort for the
manual labeling of objects or attributes, we take the advantage of
two datasets where the semantic categories are already available
in some form.
We performed our experiments on two different datasets, MIT
Indoor \cite{ref20} and SUN Attribute Dataset \cite{ref21}. MIT Indoor Dataset
is a challenging scene dataset because of its large intra-class
variations and cluttered scenes. SUN Attribute Dataset contains a
large number of images in a variety of scene categories.
MIT Indoor Dataset was an appealing option, because of its pre-
trained object models made available by Fei Fei Li et al. \cite{ref10}.
Sun Attribute Dataset meets our demands with its pre-trained 102
discriminative attribute classifiers.

\subsection{MIT Indoor Dataset}
\label{sec:sec31}
MIT Indoor dataset contains 15620 images from 67 indoor
categories. Categories do not have the same number of images,
but each category has at least 100 images. Some of these
categories are similar and hard to distinguish, such as kitchen and
restaurant kitchen, corridor and lobby etc.

For our implementation in the MIT Indoor Dataset, we split the
data into test and training sets using the recommended setup by
its authors. We use 80 images from each class for training and 20 images from each class for testing. Our implementation on this
dataset consists of using four concept groups. 

While our framework supports any number of concept groups, to
demonstrate our idea, we have only used four different types of
concept groups for the MIT Indoor Dataset. These concept groups
are well-known low-level features like PHOW \cite{ref23}, HOG \cite{ref24}, and
OPP-PHOW , besides the object detection confidence values obtained using Object Bank \cite{ref10} detectors.

\subsection{SUN Attribute Dataset}
\label{sec:sec32}
Sun Attribute Dataset is a large scale dataset consisting of more
than 700 categories and 14,000 images. 
Most of the
categories are very similar to each other with little or no
difference. This setup makes it extremely hard for scene
recognition, caused by not having enough data for each category,
and inter-class category similarity.


In order to make the scene recognition in this dataset more
feasible for our task, we have decided not to work with original
We have decided to use the images in the second-level
hierarchy category that they belong to. Some of the examples of
these categories are ``indoor sport and leisure'', ``outdoor natural
water, ice, snow'', and ``outdoor man-made sports fields, parks,
leisure spaces''. 

SUN Attribute Dataset dataset comes with a set of pre-computed
visual features that is available on its website. Among these descriptors, we used HOG, SSIM, and Attribute Confidence Vectors. Please refer to the original paper \cite{ref28} for more
details). 

\vspace{-0.3cm}
\section{Experiments}
\label{sec:sec4}
This section discusses our experiments and their results, and is
divided into two parts; self-evaluation and comparison with
others. In the first sub-section, we discuss our experiments with
different configurations of ConceptVision, and show their results.
Then in the second part, we compare the results of our method
with other methods and discuss the differences.
\vspace{-0.15cm}
\subsection{Self Evaluation}
\label{sec:sec41}

In order to evaluate ConceptVision thoroughly, see its behavior by
making slight changes and find the most efficient and simple
version of our method, we experimented with various
modifications. In this section, we discuss each modification, and
compare their results that we obtained using MIT Dataset. We
recommend reader to revisit Figure \ref{fig:method_figure} in order to have a visual
representation of our framework before reading this section.

\subsubsection{Evaluation Using Different Ensemble Techniques}
\label{sec:sec411}

This section presents the possibilities of using different ensemble
methods to combine vectors from different concept groups before
using a classifier to make a final decision.

\textbf{Confidence Summation without Weighted Classifier
Ensemble:} This version of our framework simply sums the
confidence values obtained from classifiers of different concept
groups. For the scenario of eliminating the weighted classifier
ensemble step, we treat each classifier with equal importance and
do not consider any weighting to their results. This approach can
be disadvantageous when one or more classifiers do not perform
well with the test data, and effect the final prediction in a negative
way.

\textbf{Confidence Summation with Weighted Classifier
Ensemble:} This scenario is an extended version of the
configuration presented in the first bullet of this sub-section. For
this configuration, before combining the confidence values of
each classifier in the summation step, we multiply each of them
by the corresponding class priority value, which is explained in
Section \ref{sec:sec2}. Using this approach, classifiers that are expected to
perform poorly would have less negative impact on the final
decision, and classifiers that are expected to perform well would
have more positive impact.

\textbf{Ranking without Weighted Classifier Ensemble:} To
experiment with a different ranking method we integrate a classic
ranking system \cite{ref31} to combine different features. Instead of
using exact confidence values, we sort the confidence values of
each class and rank each class in the order of preference. If we
assume there are $m$ classes, the class with the highest confidence
receives the rank $m$, while the class with the lowest confidence
receives 1. This quantizes the rank of each class to a whole
number, but some information about the classifier results can be
lost by avoiding the exact probabilities of each class. In our
experiment, instead of summing up exact confidence values; we
sum their ranks to come up with a final decision.

\textbf{Ranking with Weighted Classifier Ensemble:} This
variety of ensemble methods weighs the class ranks from
classifier by its \textit{concept-priority} value, in order to avoid the possible
issues that can rise from treating each classifier equally. As
described in \textit{confidence summation with weighting}, this
approach hopes to assign appropriate weights to the classifiers
that are expected to behave well or poorly.

\textbf{Two-Layer Classifier as Ensemble:} Another approach
to combine results of individual classifiers is using another
classifier as the ensemble method. The input of this classifier
would be the output of the previous classifiers concatenated
together. This approach has a danger of over-fitting on the second
classifier.

\begin{figure}
\centering
\includegraphics[width=0.3\textwidth]{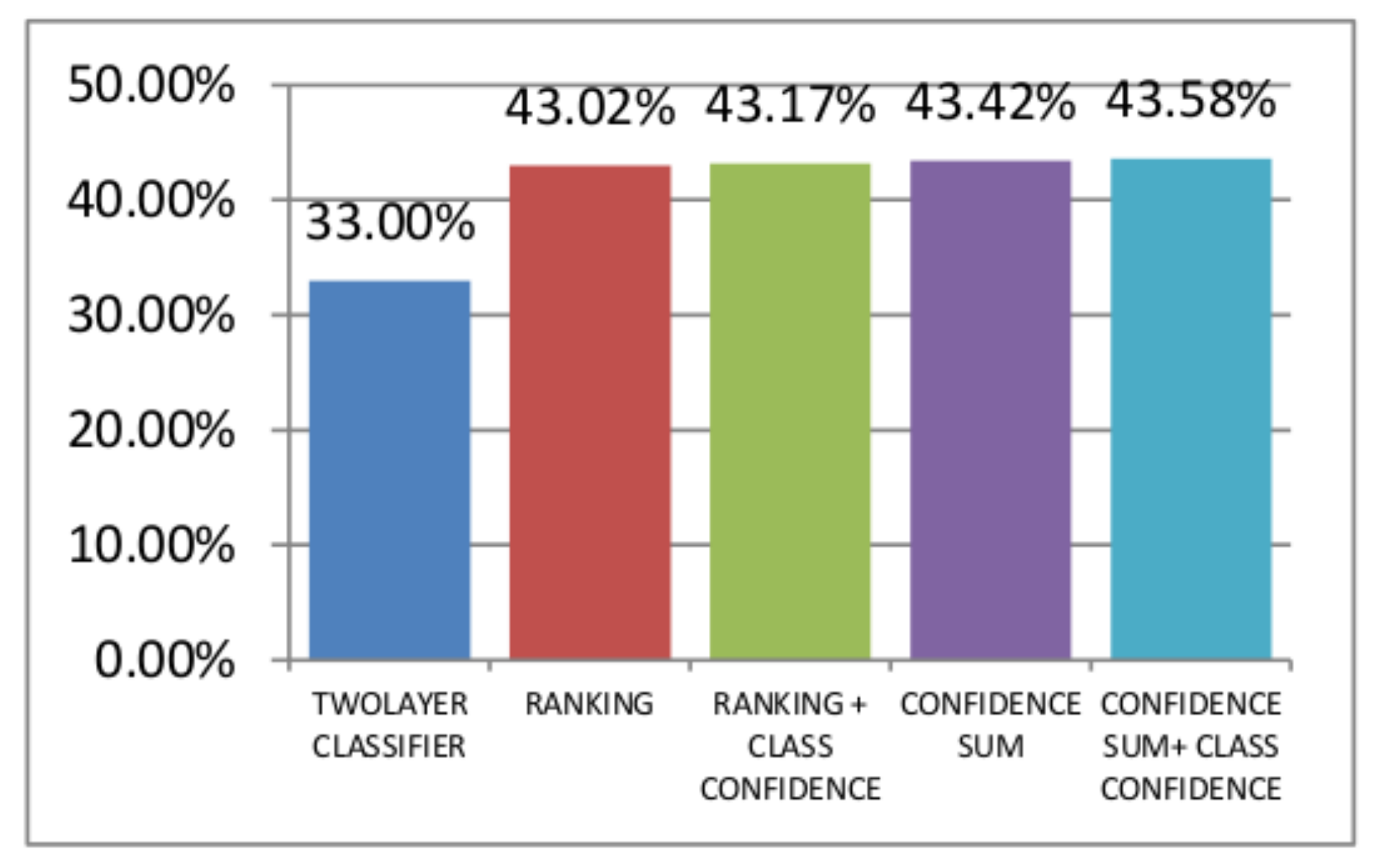}
    \vspace{-0.3cm}
\caption{The Results of Different Ensemble Techniques in
Sun Attribute Dataset}
\label{fig:fig3}
    \vspace{-0.3cm}
\end{figure}

\begin{figure}
\centering
\includegraphics[width=0.3\textwidth]{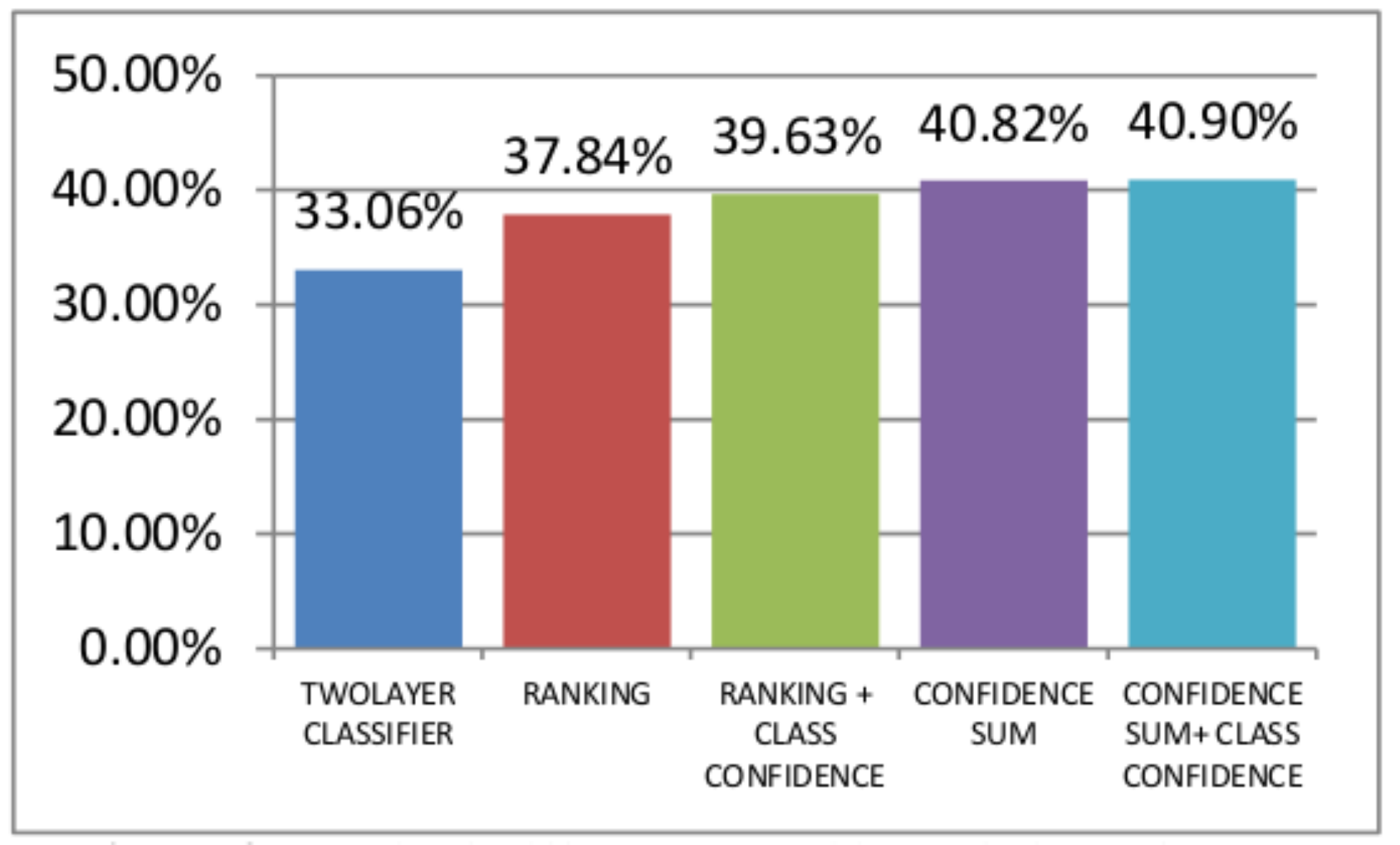}
    \vspace{-0.3cm}
\caption{The Results of Different Ensemble Techniques for MIT
Indoor Dataset}
\label{fig:fig4}
\vspace{-0.6cm}
\end{figure}

The comparison of these approaches for SUN Attributes Dataset
and MIT Indoor Dataset can be seen in Figure \ref{fig:fig3} and Figure \ref{fig:fig4}
respectively. We can see that although changing the ensemble
method did not have much effect in the Sun Attribute Dataset, the
results of the MIT Indoor Dataset are more distinct. In MIT
Indoor, ensembling concept groups using confidence summation
and weighted methods is clearly more advantageous than using a
ranking system or a non-weighted system. Using confidence-
based methods reduces the probability of losing information
classifier information, and class-priorities give each classifier
their assumed generalized performance rate. We can argue for the
same trend in SUN Attribute Dataset, but the difference of
accuracies in SUN Attribute Dataset is much less. Two-layer
Classifiers gives us the worst results for both datasets, because the
second level classifier is extremely prone to over-fitting the output
of the first layer classifier during the training stage, hence not
working well in the testing stage.

\subsubsection{Evaluation Using Different Number of
Concept Groups}
\label{sec:sec412}

\begin{figure*}[t]
\centering
\begin{tabular}{c c}
\subfigure[MIT Indoor Dataset - Weighted Confidence Method]{\includegraphics[width=0.35\linewidth]{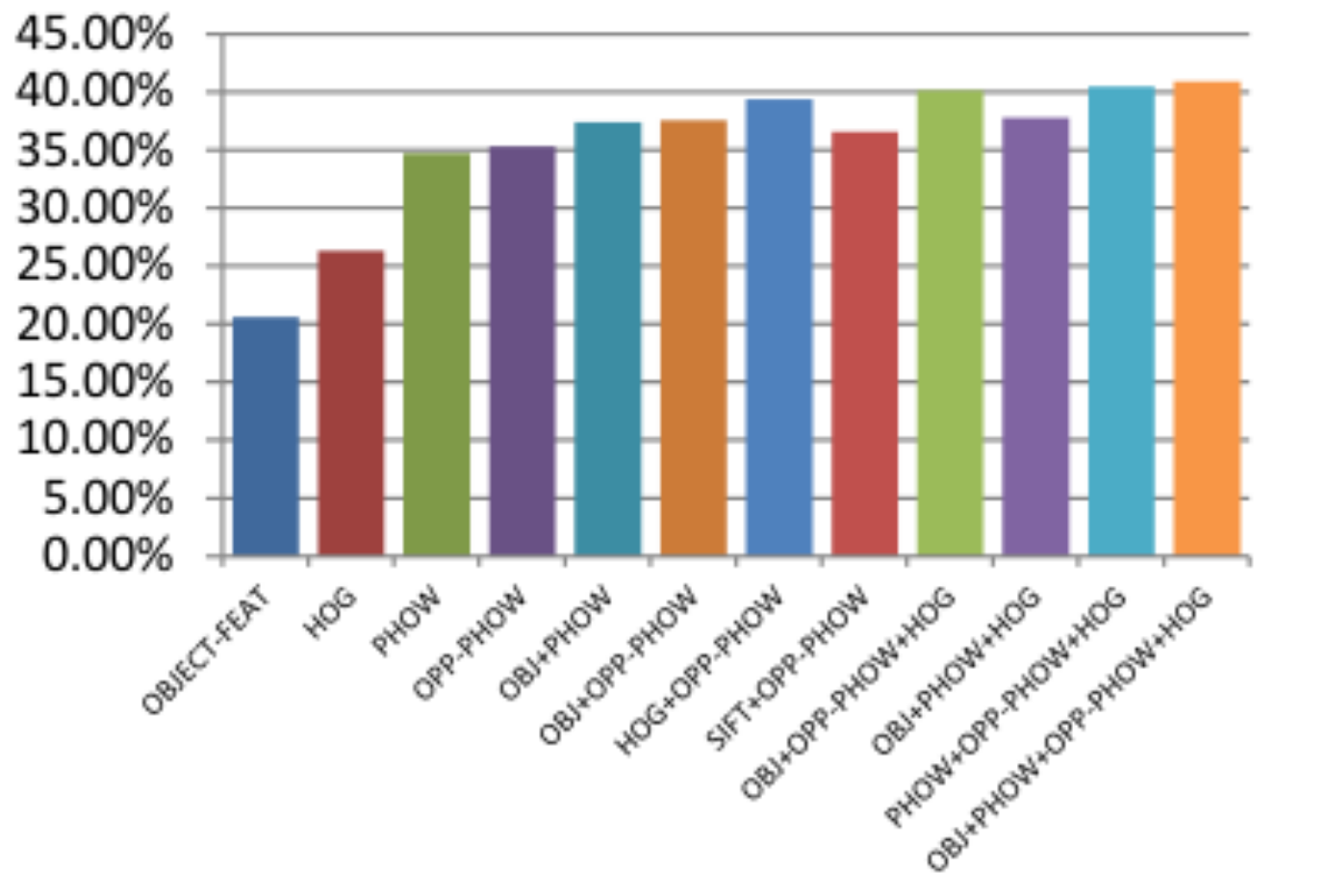}}        &
\subfigure[SUN Attribute Dataset - Weighted Confidence Method]{\includegraphics[width=0.35\linewidth]{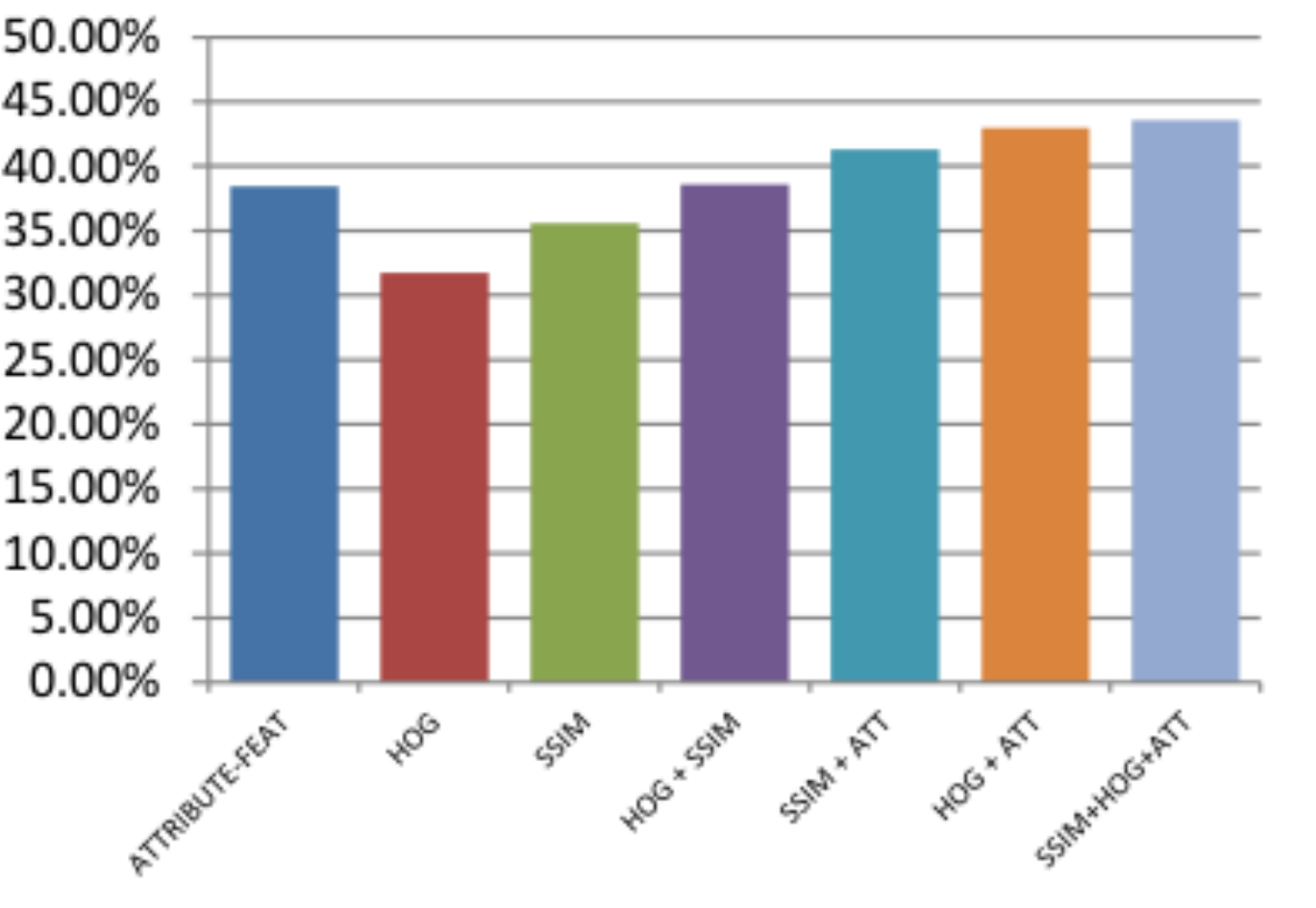}} \\
\subfigure[MIT Indoor Dataset - Weighted Ranking Method]{\includegraphics[width=0.35\linewidth]{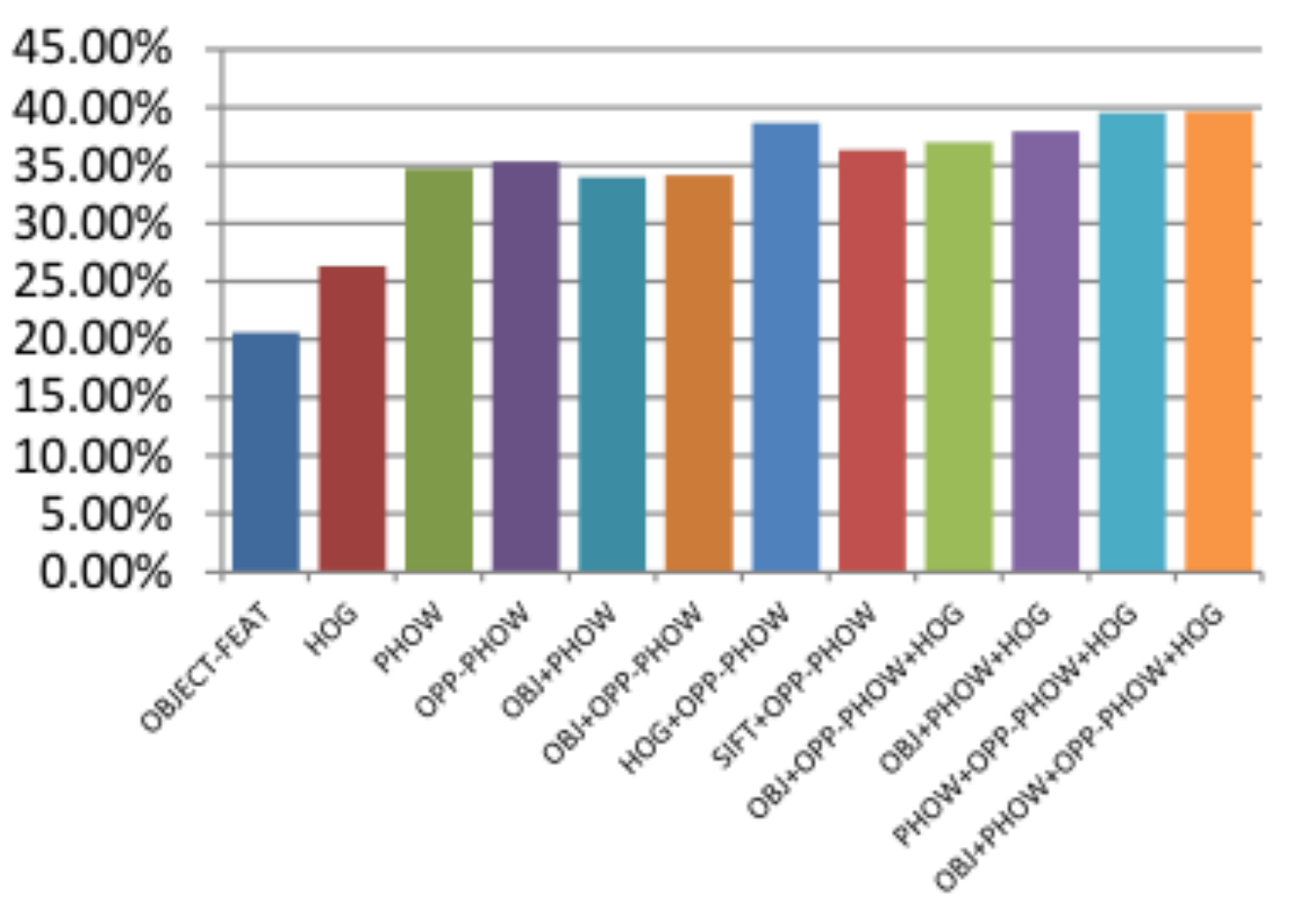}} &
\subfigure[SUN Attribute Dataset - Weighted Ranking Method]{\includegraphics[width=0.35\linewidth]{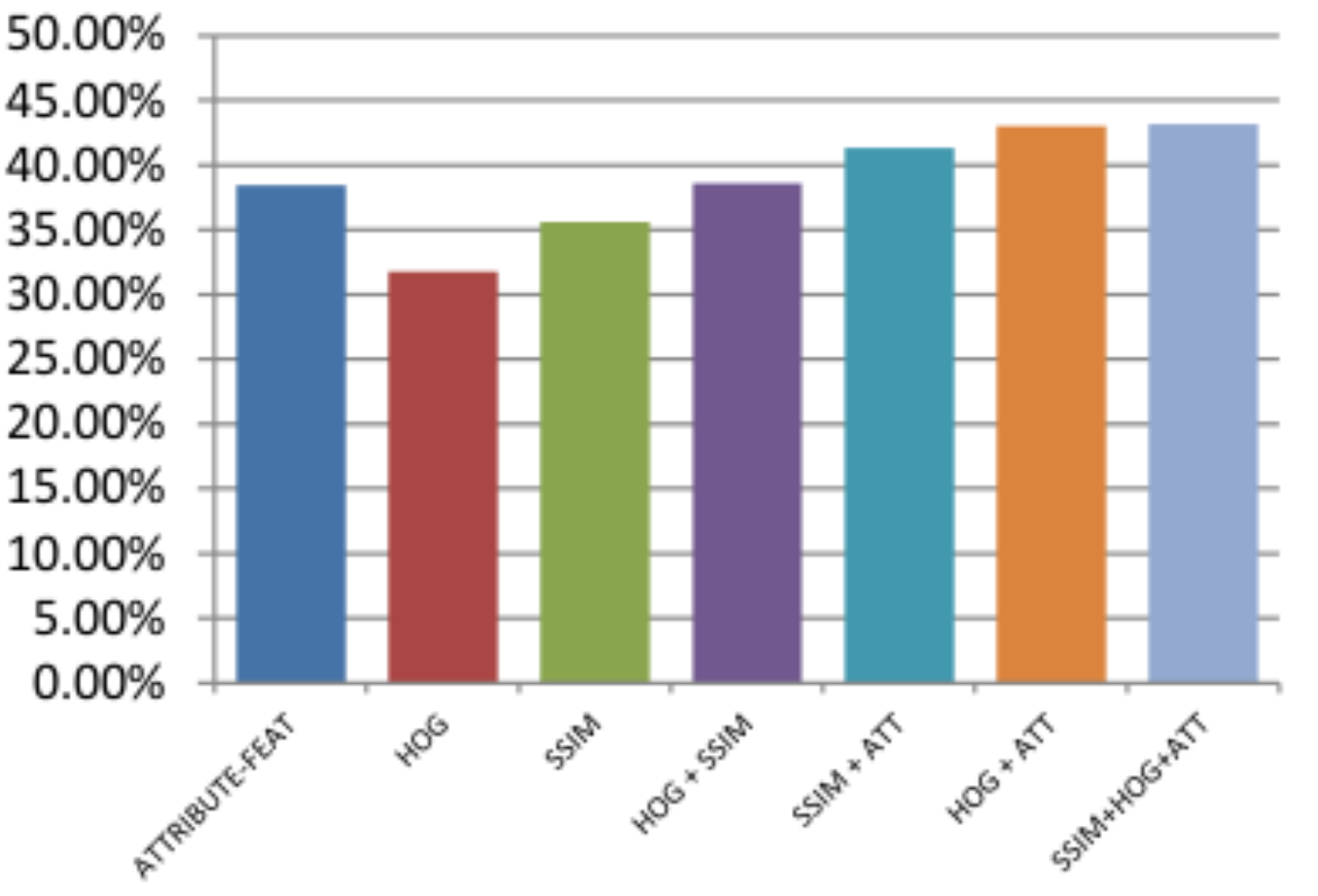}}
\end{tabular}
    \vspace{-0.25cm}
\caption{Experimental results for comparing ConceptVision with different numbers of concept groups.}
\label{fig:fig5}
    \vspace{-0.3cm}
\end{figure*}

The second different set of experiments we performed on
ConceptVision was done by changing the number of different
concepts used in each dataset. For
individual classifiers, we use probabilistic SVM, and for ensemble
of classifiers phase, we use the weighted versions of methods
discussed in Section \ref{sec:sec411}. SVM parameters are set using cross-validation on training data. Results for both datasets are reported
in Figure \ref{fig:fig5}.

Based on our experiments, we can also observe that the accuracy
of the classifier also generally increases as we add more concept
groups to our system. For both datasets, we obtain the best results
by using the highest amount of concept groups. This can serve as a proof that the combination features from completely different
concept groups can be beneficial to the overall classifier, and that
our method makes use of this relation in a meaningful way.

\subsubsection{Evaluation Using Different Classifiers}
\label{sec:sec413}

For this set of experiments, we used a fixed ensemble
configuration and changed the type of our classifier in order to
observe any different behaviors. We originally designed
ConceptVision considering the popular Support Vector Machine
classifier, however we believe it would also be necessary to see
the performance of our framework using different classifiers. To
test our framework, we also use Random Forests \cite{ref32} and Ada-
Boost \cite{ref33}.

Ada-boost is a famous ensemble method relying on the ensemble
of weak classifiers (ID3, Decision Stumps). Each weak classifier
is weighted iteratively with respect to its classification accuracy
compared to preceding ones, and they are trained with a weighted
boosting approach where the probability of an instance being
selected is proportional to its hardness of classification. 
Random Forests \cite{ref32} is a descendant of Ada-boost that also uses
the group of different Decision Trees. Each decision tree is
branched up to leaves by a randomly chosen subsample of the
training data and each node is selected by any measure that
considers randomly sampled features from the whole feature
space. 

\begin {table}
\begin{center}
 \begin{tabular}{|c|c|c|}
 \hline
 & Confidence & Ranking \\ \hline
Random Forests & 37.3\% & 32.3\% \\ \hline
Ada-Boost & 35.8\% & 33.9\% \\ \hline
SVM & 43.6\% & 43.2\% \\ \hline       
    \end{tabular}    
\caption{Comparison of different classifiers on MIT Indoor Dataset.}
\label{table11}
    \vspace{-0.7cm}
\end{center}

\end{table}
    
For these two experiments, ensemble of SVMs gives better
accuracy for the given datasets, as seen on Table \ref{table11} and \ref{table12}. These results underline
the fact that, LIBSVM's \cite{ref34} implementation of SVM
outperforms the other two classifiers with its capability of
constructing non-linear decision boundaries. Random Forests and
Ada-Boost methods are based on weak classifiers dividing the
decision space with linear boundaries.

\begin{table}

\begin{center}
\begin{tabular}{|c|c|c|}
 \hline
 & Confidence & Ranking \\ \hline
Random Forests & 32.7\% & 33.3\% \\ \hline
Ada-Boost & 33.2\% & 34.7\% \\ \hline
SVM & 40.9\% & 39.6\% \\ \hline       
\end{tabular}
\caption{Comparison of different classifiers on SUN Attribute Dataset.} 
\label{table12}   
    \end{center}
    \vspace{-0.7cm}

\end{table}

\vspace{-0.2cm}
\subsection{Comparison with Other Studies}
\label{sec:sec42}

In this section we compare the results of ConceptVision from
Section \ref{sec:sec41} with some of the other methods used in literature.
Since the Sun Attribute Dataset is relatively recent and there are
not many experimental results with it in the literature, we make
our comparisons using the results obtained from MIT Indoor
Dataset. You can see the comparisons on Table \ref{table2}, and see the
detailed explanations of the compared methods in the following
sub-sections.

\begin{table}

\begin{center}
\begin{tabular}{|c|c|}
 \hline
  Method & Accuracy \\ \hline
Feature Concatenation & 9.48\% \\ \hline
OB-LR \cite{ref10} & 37.6\% \\ \hline
ConceptVISION & 40.9\% \\ \hline       
\end{tabular}
\caption{Comparison of ConceptVision with feature
concatenation and Object Bank \cite{ref10} in MIT Indoor Dataset.} 
\label{table2}   
    \end{center}
    \vspace{-0.8cm}
\end{table}

For comparing with other methods, we designate the feature 
concatenation method as our baseline, and also compare 
ConceptVision with Object Bank \cite{ref10}, which we consider the 
state-of-art. 
\vspace{-0.25cm}
\subsubsection{Object Bank}
\label{sec:sec421}
Object Bank \cite{ref10} is a well known
method with the idea of having a higher semantic level description of images, exposing scene's semantic structure similar to human
understanding of views. Although ObjectBank provides a good interpretation of the image, it produces a very high dimensional vectors, such as 42588 dimensions with the configuration in the original paper.



\subsubsection{Feature Concatenation}
\label{sec:sec422}

Lastly, classical method to combine different concepts or features
is to just concatenate them horizontally. This method is extremely
simple and widely used, but it can have many disadvantages, such
as resulting features being in very high dimensions. Also,
combining features from very different concepts, such as low-
level and high-level features, does not necessarily add any
meaning for classification purposes, and can provide low results.

\subsubsection{Results Comparison}
\label{sec:sec423}

When comparing our results with those of Object Bank and
feature concatenation on MIT Indoor Dataset, we can see that our
method performs better than Object Bank. Results of the
comparison can be seen Table 2. The results of feature concatenation performing poorly are not surprising, as we had
expected a large concatenation of features from different levels
not to perform well. In the second comparison, our method
performs actually better than Object Bank on this dataset, without
having an as complex object representation as Object Bank does.
This shows us that, when combining different concept groups
from different levels, their resulting accuracy can be higher than
using one of the concepts groups.
\vspace{-0.3cm}
\section{Discussion and Feature Work}
\label{sec:sec5}

We proposed ConceptVision as a framework for combining
concept groups from many different levels and perspective for the
purpose of scene categorization. The proposed framework
provides flexibility for supporting any type of concept groups,
such as those that have semantic meanings like objects and
attributes, or low-level features that have no meanings
semantically but can provide important information about the structure of an image. There is no limit in the definition of
concepts, and it is easy to be expanded through inclusion of any
other intermediate representation describing the whole or part of
the image in content or semantics.

Although our work can be further improved in many ways, the
experimental results from Section \ref{sec:sec4} look very promising. We
show that individual concept group classification accuracies are
lower compared to their combination with ConceptVision
framework. Furthermore, as we improve the number of concept
groups that are used, we also obtain higher accuracies. This can be
seen especially in the MIT dataset, where even a concept group
like the object representation, which has a poor accuracy overall,
increases the overall accuracy when combined with the other low-
level features.

Another conclusion we can derive from our experiments is that
weighting the results of each individual classifier by its class-
priority value also improves the overall accuracy. This can also
be seen by our experiments, where the weighted versions of both
ensemble methods provide better results than their non-weighted
version for both datasets.

%
%
Although we are encouraged by its results, we must also express
that ConceptVision is far from perfect and there are many
improvements that can be made on top of it. Our framework
examines each concept group on the same level, by assuming that
their classification models are completely independent from each
other. We can extend our framework by modifying this idea, and
establishing dependence between each concept group by their
semantic meanings. This way, we would still be able to
incorporate all kinds of concepts groups from different levels, but
we would use them in different hierarchical levels in more
meaningful way.
\vspace{-0.3cm}

%
\bibliographystyle{ieeetr}
\bibliography{paper}  

\begin{thebibliography}{10}

\bibitem{ref1}
M.~Szummer and R.~W. Picard, ``Indoor-outdoor image classification,'' in {\em
  In IEEE Intl. Workshop on Content-Based Access of Image and Video Databases},
  1998.

\bibitem{ref2}
A.~Vailaya, A.~Member, M.~A.~T. Figueiredo, A.~K. Jain, H.-J. Zhang, and
  S.~Member, ``Image classification for content-based indexing,'' {\em IEEE
  Transactions on Image Processing}, vol.~10, pp.~117--130, 2001.

\bibitem{ref3}
A.~Payne and S.~Singh, ``Indoor vs. outdoor scene classification in digital
  photographs,'' {\em Pattern Recogn.}, vol.~38, pp.~1533--1545, Oct. 2005.

\bibitem{ref4}
N.~Serrano, A.~E. Savakis, and J.~Luo, ``A computationally efficient approach
  to indoor/outdoor scene classification.,'' in {\em ICPR (4)}, 2002.

\bibitem{ref5}
A.~Oliva and A.~Torralba, ``Modeling the shape of the scene: A holistic
  representation of the spatial envelope,'' {\em Int. J. Comput. Vision},
  vol.~42, pp.~145--175, May 2001.

\bibitem{ref6}
P.~Quelhas, F.~Monay, J.~M. Odobez, D.~Gatica-Perez, and T.~Tuytelaars, ``A
  thousand words in a scene,'' {\em IEEE Trans. Pattern Anal. Mach. Intell.},
  vol.~29, pp.~1575--1589, Sept. 2007.

\bibitem{ref7}
F.-F. Li and P.~Perona, ``A bayesian hierarchical model for learning natural
  scene categories,'' in {\em CVPR}, 2005.

\bibitem{ref8}
A.~Bosch, A.~Zisserman, and X.~Mu\~{n}oz, ``Scene classification using a hybrid
  generative/discriminative approach,'' {\em IEEE Trans. Pattern Anal. Mach.
  Intell.}, vol.~30, pp.~712--727, Apr. 2008.

\bibitem{ref9}
S.~Lazebnik, C.~Schmid, and J.~Ponce, ``Beyond bags of features: Spatial
  pyramid matching for recognizing natural scene categories,'' in {\em CVPR},
  2006.

\bibitem{ref10}
E.~P.~X. Li-Jia~Li, Hao~Su and L.~Fei-Fei, ``Object bank: A high-level image
  representation for scene classification \& semantic feature sparsification,''
  in {\em NIPS}, 2010.

\bibitem{ref11}
L.-J. Li, H.~Su, Y.~Lim, and L.~Fei-Fei, ``Objects as attributes for scene
  classification,'' in {\em ECCV}, 2012.

\bibitem{ref12}
A.~Farhadi, I.~Endres, D.~Hoiem, and D.~Forsyth, ``Describing objects by their
  attributes,'' in {\em CVPR}, 2009.

\bibitem{ref13}
C.~H. Lampert, H.~Nickisch, and S.~Harmeling, ``Learning to detect unseen
  object classes by betweenclass attribute transfer,'' in {\em In CVPR}, 2009.

\bibitem{ref14}
L.~Torresani, M.~Szummer, and A.~Fitzgibbon, ``Efficient object category
  recognition using classemes,'' in {\em ECCV'10}, 2010.

\bibitem{ref15}
J.~Vogel and B.~Schiele, ``Semantic modeling of natural scenes for
  content-based image retrieval,'' {\em Int. J. Comput. Vision}, vol.~72,
  pp.~133--157, Apr. 2007.

\bibitem{ref16}
A.~Hauptmann, R.~Yan, W.-H. Lin, M.~Christel, and H.~D. Wactlar, ``Can
  high-level concepts fill the semantic gap in video retrieval? a case study
  with broadcast news.,'' {\em IEEE Transactions on Multimedia}, vol.~9, no.~5,
  pp.~958--966, 2007.

\bibitem{ref17}
J.~Smith, M.~Naphade, and A.~Natsev, ``Multimedia semantic indexing using model
  vectors,'' in {\em ICME}, 2003.

\bibitem{ref20}
A.~Quattoni and A.~Torralba, ``Recognizing indoor scenes,'' 2009.

\bibitem{ref21}
G.~Patterson, ``Sun attribute database: Discovering, annotating, and
  recognizing scene attributes,'' in {\em CVPR}, 2012.

\bibitem{ref23}
A.~Bosch, A.~Zisserman, and X.~Mu\~noz, ``Image classification using random
  forests and ferns,'' in {\em ICCV}, 2007.

\bibitem{ref24}
N.~Dalal and B.~Triggs, ``Histograms of oriented gradients for human
  detection,'' in {\em CVPR}, 2005.

\bibitem{ref28}
J.~Xiao, J.~Hays, K.~A. Ehinger, A.~Oliva, and A.~Torralba, ``Sun database:
  Large-scale scene recognition from abbey to zoo,'' in {\em CVPR}, 2010.

\bibitem{ref31}
T.~K. Ho, J.~J. Hull, and S.~N. Srihari, ``Decision combination in multiple
  classifier systems,'' {\em IEEE Trans. Pattern Anal. Mach. Intell.}, vol.~16,
  pp.~66--75, Jan. 1994.

\bibitem{ref32}
L.~Breiman, ``Random forests,'' {\em Mach. Learn.}, vol.~45, pp.~5--32, Oct.
  2001.

\bibitem{ref33}
Y.~Freund and R.~E. Schapire, ``A decision-theoretic generalization of on-line
  learning and an application to boosting,'' in {\em EuroCOLT}, 1995.

\bibitem{ref34}
C.-C. Chang and C.-J. Lin, ``{LIBSVM}: A library for support vector machines,''
  {\em ACM Transactions on Intelligent Systems and Technology}, vol.~2,
  pp.~27:1--27:27, 2011.

\end{thebibliography}
%
%
\end{document}